\documentclass[10pt,twocolumn,letterpaper]{article}

\usepackage{3dv}
\usepackage{times}
\usepackage{epsfig}
\usepackage{graphicx}
\usepackage{amsmath}
\usepackage{amssymb}
\usepackage{gensymb}
\usepackage{pifont}
\newcommand{\cmark}{\ding{51}}
\newcommand{\xmark}{\ding{55}}

\usepackage[pagebackref=true,breaklinks=true,letterpaper=true,colorlinks,bookmarks=false]{hyperref}

\threedvfinalcopy 

\def\threedvPaperID{****} 

\ifthreedvfinal\fi
\begin{document}

\title{PanoNet3D: Combining Semantic and Geometric Understanding for LiDAR Point Cloud Detection}

\author{Xia Chen, Jianren Wang, David Held, Martial Hebert\\
Robotics Institute, Carnegie Mellon University\\
5000 Forbes Ave, Pittsburgh, Pennsylvania\\
{\tt\small \{xiac, jianrenw, dheld, hebert\}@cs.cmu.edu}
}

\maketitle

\begin{abstract}

Visual data in autonomous driving perception, such as camera image and LiDAR point cloud, can be interpreted as a mixture of two aspects: semantic feature and geometric structure. Semantics come from the appearance and context of objects to the sensor, while geometric structure is the actual 3D shape of point clouds. Most detectors on LiDAR point clouds focus only on analyzing the geometric structure of objects in real 3D space. Unlike previous works, we propose to learn both semantic feature and geometric structure via a unified multi-view framework. Our method exploits the nature of LiDAR scans -- 2D range images, and applies well-studied 2D convolutions to extract semantic features. By fusing semantic and geometric features, our method outperforms state-of-the-art approaches in all categories by a large margin. The methodology of combining semantic and geometric features provides a unique perspective of looking at the problems in real-world 3D point cloud detection.
\end{abstract}

\section{Introduction}
With the recent advent of autonomous vehicles, detecting and localizing obstacles on LiDAR point clouds has become a popular research topic. While the output of LiDAR sensors is three-dimensional, it is fundamentally different than true 3D data (such as 3D mesh models). Because of the sweeping mechanics of LiDAR, the data can be densely represented in 2D format (range image). This is commonly referred to as 2.5D \cite{Goldstein1988Two-DimensionalUnwrapping}. Many popular 3D detectors like PointPillars~\cite{Lang2018PointPillars:Clouds} often ignore such fact and treat the LiDAR data purely as a collection of $(x,y,z,i)$ points ($i$ is the point's intensity or reflectance). Though these works achieve good performance on detection tasks, they do not take advantage of the intrinsic structure of the data. 

The simplest way to address this issue is to format the data as normal 2D images and to apply well-studied 2D image detectors on them. However, this solution has several drawbacks. First, spatial coordinates are fundamentally different from images' RGB values. The spatial structure of points cannot be easily extracted from 2D convolutions on the projected range image. Second, range images are not scale-invariant. That is, closer objects contain a much larger number of pixels compared to objects that are far away. Various scales of objects make it hard for the network to generalize. 

\begin{figure}
    \centering
    \includegraphics[width=0.9\linewidth]{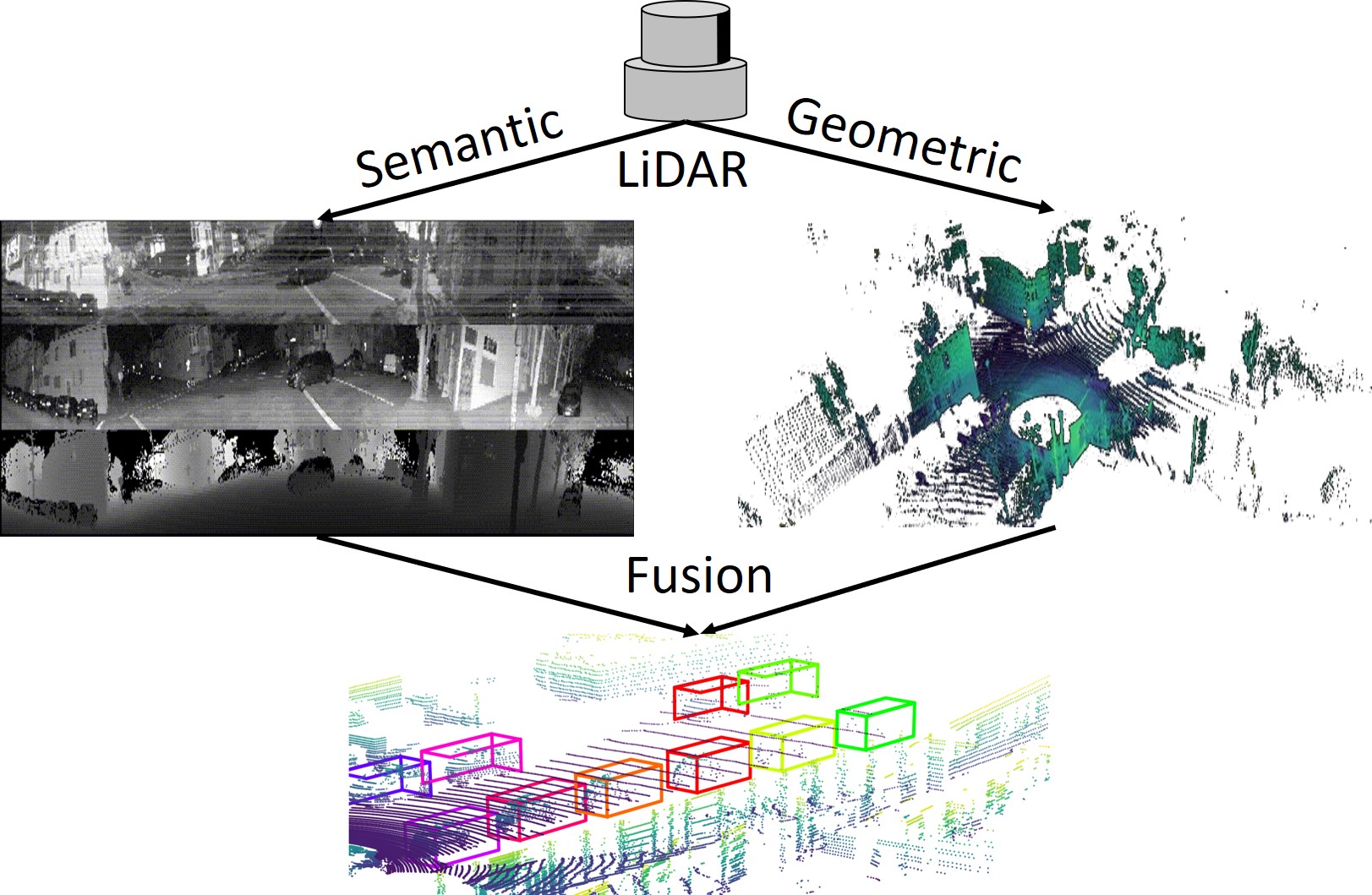}
    \caption{LiDAR can be interpreted semantically and geometrically by nature. \textit{PanoNet3D} utilize both information for LiDAR object detection.}
    \label{fig:teaser}
\end{figure}

While 2D convolution is not efficient at understanding the 3D geometric structure of point clouds, it can still extract meaningful deep semantic features from range images just like from conventional color pictures. We argue that combining both deep semantic features from range images and raw geometric structures from 3D point clouds together can yield better detection results. More specifically, in the first step, we extract semantics from the projected range image with a fully convolutional network (FCN). The output high dimensional semantic features are then fused with low dimensional raw geometric features which are usually computed by simple geometric manipulations or shallow networks. Final predictions are generated from a main-stage detector with a 3D sparse convolutional network as the backbone. In such manner, we utilize semantics from 2.5D range images while keeping scale invariance in 3D space at the same time. Our experiment shows that additional semantic features significantly improve the detection performances on NuScenes~\cite{Caesar2019NuScenes:Driving} dataset, surpassing the current first-place method CBGS~\cite{Zhu2019Class-balancedDetection} on the official leader-board. The key contributions of this work are the following:

\begin{itemize}
  \item We introduce PanoNet3D, a novel approach that feeds both deep semantic features and raw geometric features of point cloud data to the main detector. By doing so, the detector is exposed to both the spatial structure of point cloud as well as semantic information natural to the LiDAR sensor. 
  \item PanoNet3D achieves significant improvements on both single-sweep input and multiple-sweep LiDAR input. Our design of temporal aggregation allows aggregating multiple scanned frames for denser input data without the redundancy of repeatedly running the same semantic feature extraction network on these frames.
  \item The integration of a pano-view feature extractor of PanoNet3D enables natural and simple removal of occluded points after a crop-and-paste data augmentation. Handling occlusions of augmented objects is hard in bird-eye view and is often ignored by BEV detectors.
  \item PanoNet3D beats state-of-the-art (SOTA) performance on 3D object detection. With several improvements on network architectures, it achieves 0.54 mAP on NuScenes dataset detection challenge, out-performing PointPillars \cite{Lang2018PointPillars:Clouds} and CBGS \cite{Zhu2019Class-balancedDetection}.

\end{itemize}   

\section{Related Work}

\begin{figure*}[h]
\centering
  \includegraphics[width=0.85\linewidth]{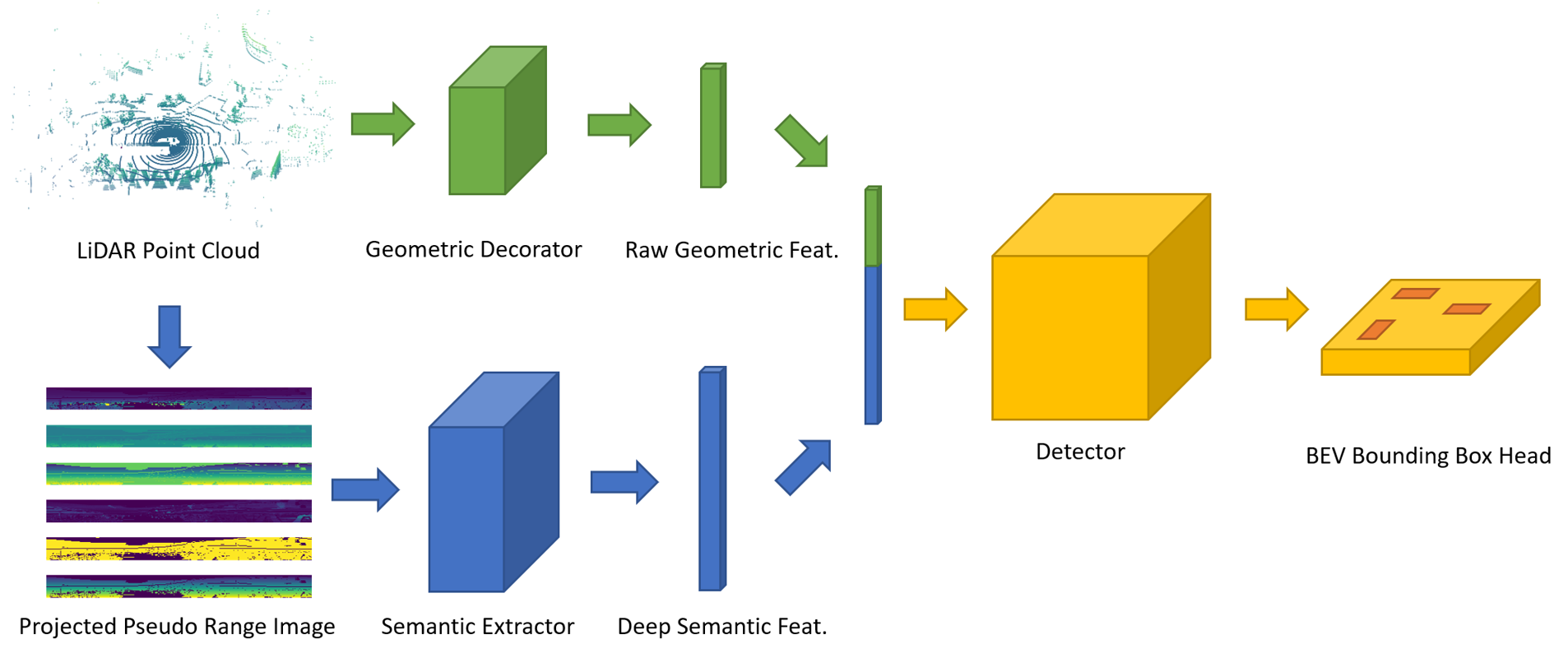}
\caption[PanoNet3D's framework for point cloud detection]{PanoNet3D's framework for point cloud detection. The top branch takes LiDAR point cloud as input and decorates raw point features with several simple local geometric features. The lower branch converts point cloud to pseudo range image and feeds it into a 2D FCN to get per-pixel deep semantic feature. The output features of these two branches are then aggregated and passed to the main detector. A final bounding box head generates detected proposals on the BEV plane.}
  \label{fig:panonet3d_framework}
\end{figure*}

\subsection{Point Cloud Representation}

Deep learning architectures take different formats of 3D point clouds as input. The first class consumes raw point clouds directly, including PointNet~\cite{Qi2017PointNet:Segmentation}, PointNet++~\cite{Qi2017PointNet++:Space}, and PointRCNN~\cite{Shi2018PointRCNN:Cloud}. This type of approaches require no pre-processing of point clouds (such as voxelization or rendering), but their performance suffers when the scene is large and sparse. For common LiDAR sensors, a single sweep typically contains over 50,000 points. So these networks usually need to down-sample input, losing the resolution of raw data. 

Some networks simply treat point clouds as a bird-eye-view (BEV) image, {\em e.g.}, AVOD~\cite{Ku2018JointAggregation} and  Complex-YOLO~\cite{Simon2018Complex-YOLO:Clouds}. BEV images work particularly well for LiDAR point clouds as we usually only care about x-y (2D) localization of objects. This formatting allows 2D image detection frameworks to be re-applied on point clouds at the cost of partly losing vertical geometric structure information. 

Another type of 3D point cloud formatting is voxelization. Examples include VoxelNet~\cite{Zhou2018VoxelNet:Detection}, SECOND~\cite{Yan2018Second:Detection}, and PIXOR~\cite{Yang2018PIXOR:Clouds}. Voxelized point cloud usually has a finite spatial size with pooling as the technique to convert per-point features to per-voxel features. The performance of this class of detectors is usually linked to voxel resolution.

Recently, LaserNet~\cite{Meyer2019LaserNet:Driving} shows that when the size of the training dataset is large enough, detectors performing on the perspective view of point clouds (range images) can achieve performance on par with BEV detectors. Similarly, MVF~\cite{Zhou2019End-to-EndClouds} extracts semantics from both range images and BEV images with two 2D convolutional towers. For point clouds scanned by LiDARs, the range image format is much denser and has no range limits compared to BEV-based representations.

\subsection{Object Detection}
Object detection has traditionally been studied on 2D images. Various Convolutional Neural Network (CNN) detectors are proposed since R-CNN~\cite{Girshick2014RichSegmentation}. These detectors can be categorized into two major classes: two-stage detectors and single-stage detectors. Two-stage detectors usually consist of a Region Proposal Network (RPN)~\cite{Ren2015FasterNetworks} that produces candidate region proposals and a second stage network regressing the final bounding boxes. On the other hand, single-stage detectors rely on a Single Shot Detector (SSD)~\cite{Liu2016Ssd:Detector} that densely produces bounding box predictions with a single fully convolutional network (FCN). Single-stage detectors are simpler and typically faster than two-stage detectors. With focal loss~\cite{Lin2017FocalDetection} alleviating the problem of foreground-background class imbalance, singlnie-stage detectors can achieve similar or even better results compared to two-stage detectors.

Object detection on 3D point clouds is a more recent research topic. Many works borrow ideas from 2D image detectors as there is no fundamental difference between these two tasks. The only necessary modification of the detection head is the regression of additional parameters required to define 3D bounding boxes. Many modern point cloud detectors adopt single-stage frameworks, including SECOND~\cite{Yan2018Second:Detection}, PointPillars~\cite{Lang2018PointPillars:Clouds}, PIXOR~\cite{Yang2018PIXOR:Clouds}, and LaserNet~\cite{Meyer2019LaserNet:Driving}. Single-stage point cloud detector is more favorable for autonomous driving applications due to its simplicity and fast inference speed.

\subsection{Detection on LiDAR Point Cloud}
Object detection on LiDAR point cloud data has several domain-specific problems. We discuss convolution types, temporal aggregation, and data augmentation below.

\subsubsection{Convolution Types}
Intuitively, voxelized point cloud data is a 3D tensor and thus the detector should consist of 3D convolution layers. Because of the sparsity of LiDAR data, GPU-accelerated sparse implementation of 3D convolution is usually applied~\cite{Yan2018Second:Detection} in order to significantly reduce time and memory consumption. PointPillars~\cite{Lang2018PointPillars:Clouds} converts 3D inputs to 2D by using a pillar feature encoder that outputs per grid feature embedding on the x-y (BEV) plane. This allows the detector to use regular 2D convolutional layers that are highly optimized on GPUs by many deep learning libraries.

\subsubsection{Temporal Aggregation}
Some detectors aggregate multiple consecutive LiDAR sweeps and show that temporal information can improve detection performance. FaF~\cite{Luo2018FastNet} treats temporal information as an additional dimension of the input tensor, i.e., multiple frames are appended along a new dimension to create a 4D tensor. SECOND~\cite{Yan2018Second:Detection} proposes a simpler solution that adds relative temporal stamp to each point as an extra input channel (the input tensor remains 3D). We need to pay special attention to ego motion during temporal aggregation as the reference coordinate system shifts with the ego vehicle's movement.

\subsubsection{Multi-view Aggregation}
MV3D~\cite{Chen2016Multi-ViewDriving} proposed a multi-view detection network which has two detection branches, one for BEV and one for range view (RV). The results of the two branches are fused afterwards. While MV3D explores the possibility of jointly using both BEV and RV for point cloud detection, the paper does not give the justification of why the two views should be used jointly. RV is what the sensor sees in raw, from which we can effectively extract semantic features just like from RGB camera images. On the other hand, BEV is scale-invariant regardless of the distance to the sensor, so actual geometric structures are preserved in BEV. The need of using both semantic and geometric information leads to the combination of RV and BEV.

\subsubsection{Data Augmentation}
Data augmentation is extremely important for training LiDAR detection networks in autonomous driving scenarios, as real-world datasets usually have severe problem of class imbalance. For example, about half of labeled instances in NuScenes~\cite{Caesar2019NuScenes:Driving} dataset are cars. A copy-and-paste augmentation schematic are used in many popular detectors including SECOND~\cite{Yan2018Second:Detection}, PointPillars~\cite{Lang2018PointPillars:Clouds}, and CBGS~\cite{Zhu2019Class-balancedDetection}. This method crops ground truth bounding boxes from other frames and pastes them onto the current frame's ground plane. Hu et al.~\cite{Hu2019WhatDetection} argue that maintaining correct visibility during augmentation makes significant improvements in detection results. The visibility information can be calculated and explicitly expressed. However, with projection on range images, visibility is naturally encoded and requires less computation.

\section{Method}
\begin{figure*}[h!]
\centering
  \includegraphics[width=0.85\linewidth]{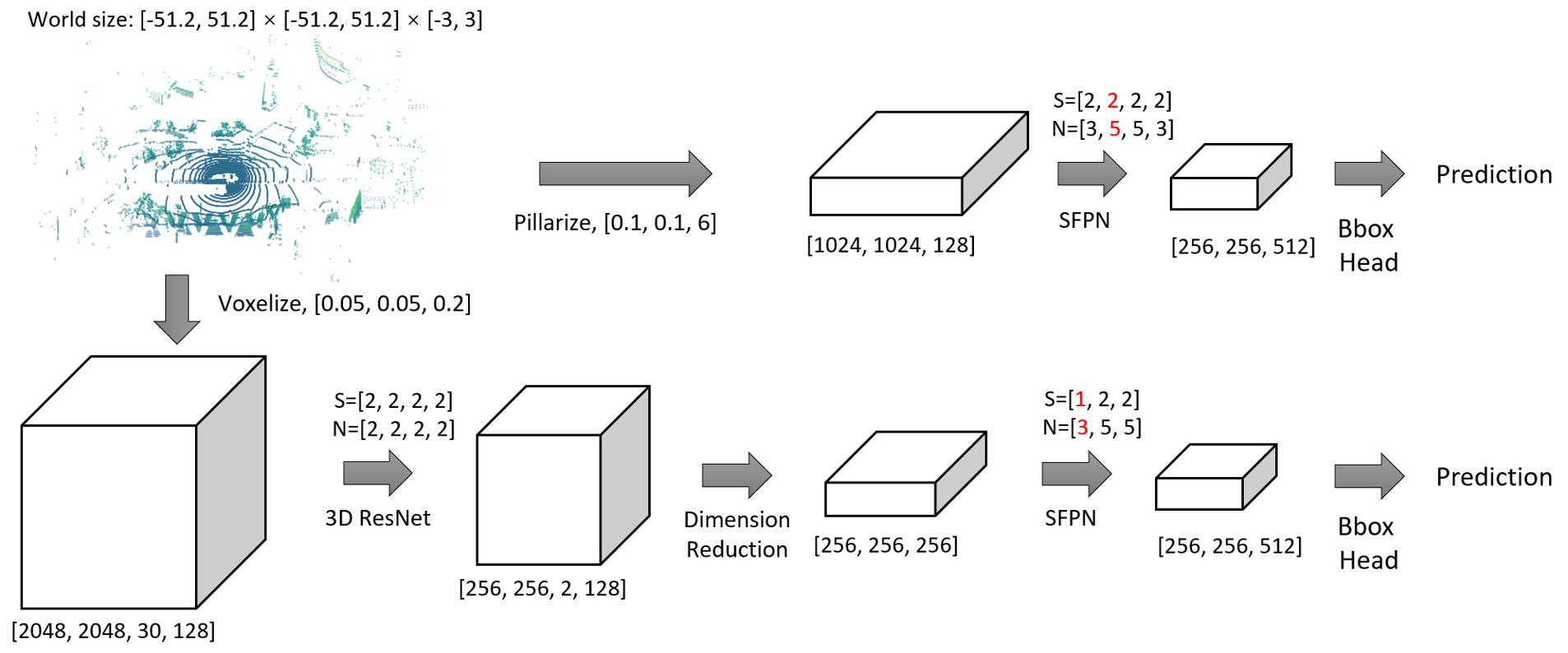}
    \caption[Structure of detector with 2D pillars or 3D voxels as input]%
    {Structure of a detector with 2D pillars or 3D voxels as input. The initial feature is 128 dimensional. We limit the size of the whole scene to $[-51.2, 51.2] \times [-51.2, 51.2] \times [-3, 3]$ meters in $x,y,z$ direction. The networks consist of a few layers of ResNet basic blocks. S denotes the stride of each layer and N denotes the number of blocks. The generated feature map of SFPN has the same resolution as the layer marked in red. }
  \label{fig:panonet3d_detector}
\end{figure*}

The structure of PanoNet3D is illustrated in Fig. \ref{fig:panonet3d_framework}. This framework can be divided into two stages. 1) \textbf{Feature extraction stage}: A 2D FCN generates deep semantic feature maps from projected pseudo range images. Meanwhile, a geometric decorator generates each point's raw geometric features, including its global position and local displacement relative to the center of its residing voxel. The semantic features and geometric features are then aggregated and passed to the next stage. 2) \textbf{Detection stage}: Per-point features are converted to per-voxel features by a simple symmetric operation such as max and average pooling. A single-stage detector then predicts oriented 3D boxes and their confidence score based on pre-defined anchors. We describe the details of each component of the network in the following sections. First, we will introduce pseudo-range-image semantic extractor and voxel geometric decorator, the combination of which generates a feature vector for each point. Then we will discuss how to temporally aggregates features from multiple frames. Last, we will describe how the main detector gives prediction as well as training and implementation details.

\subsection{Pseudo Range Image and   Semantic Extractor}

\begin{figure}[h!]
  \includegraphics[width=\linewidth]{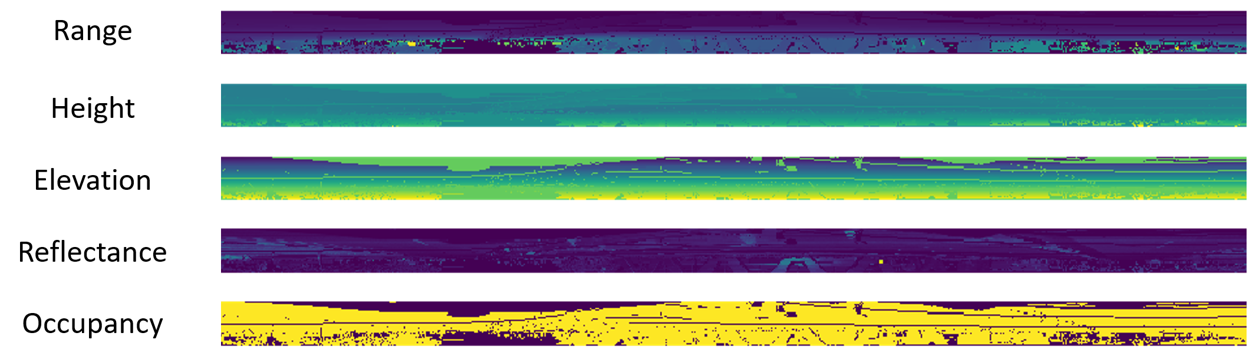}
\caption[An example of projected pseudo range image with five channels]{An example of projected pseudo range image with five channels. From top to bottom: range $r$, height $h$, elevation angle $\phi$, reflectance $i$, and occupancy mask $m$.}
  \label{fig:pseudo_range_image}
\end{figure}

The outputs of a common LiDAR sensor are range images by nature. However, since many LiDARs' rings are not evenly spaced (sometimes the ring information is not even available), we manually project 3D point clouds back to 2D range images with evenly spaced projection angles. For the NuScenes~\cite{Caesar2019NuScenes:Driving} dataset, we choose the horizontal projection angle range and resolution to be $[x_{min}, x_{max}, x_{step}]=[-180\degree, 180\degree, 0.3125\degree]$ and vertical counterparts to be $[y_{min}, y_{max}, y_{step}]=[-30\degree, 10\degree, 1.25\degree]$. It is possible that more than one LiDAR point is mapped to the same pixel on range image. In this case, we simply keep the closest point and discard the rest. In addition to point's range $r$, we also encode height $h$, elevation angle $\phi$ and reflectance $i$ in separate channels. Similar to LaserNet~\cite{Meyer2019LaserNet:Driving}, the last channel of the image is a flag indicating whether
a pixel contains a projected point. We call this multi-channel tensor (an example is shown in Fig. \ref{fig:pseudo_range_image}) \textbf{pseudo range image} of LiDAR. 

To extract semantic features from the pseudo range image, we adopt the  Semantic FPN (SFPN) design from \cite{Kirillov2019PanopticNetworks}. It aggregates the features from all levels of FPN layers into a single output with per-pixel semantic embedding. The SFPN's backbone is a ResNet34~\cite{He2016DeepRecognition} without the first layer (conv1). For each projected LiDAR points, the SFPN generates a 64-dimensional semantic feature vector. The feature extractor is not trained with direct supervision. Instead, the feature vectors are passed to the main detector where they receive supervision from the final classification and localization loss. 

\begin{figure*}[h]
  \includegraphics[width=\linewidth]{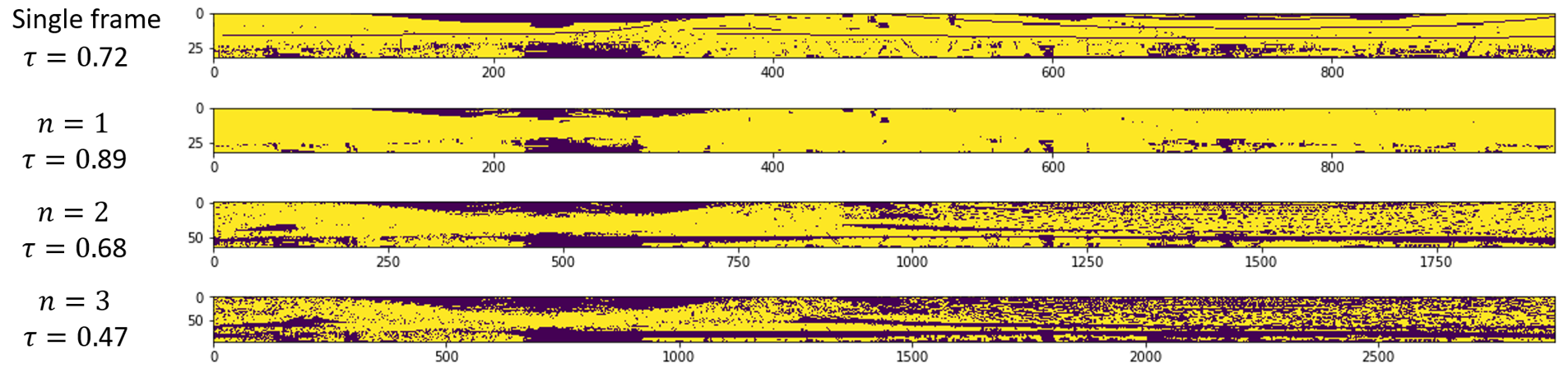}
\caption[Occupancy maps showing the results of spatial multi-frame fusion]{Occupancy maps showing the results of spatial multi-frame fusion. Yellow color indicates the pixel is occupied by a projected LiDAR point. $\tau$ is the occupancy rate (the number of occupied pixels over the number of total pixels). From top to bottom: original single frame, 10-frame aggregation with $n=1$, $n=2$, $n=3$.}
  \label{fig:occupancy_map}
\end{figure*}

\subsection{Voxelization and Geometric Decorator}  \label{secVoxel}

The input 3D point cloud is voxelized before being passed into the detector. We experiment with two types of voxelization: (1) regular 3D voxelization and (2) pillarization, where points are organized in vertical columns similar to PointPillars~\cite{Lang2018PointPillars:Clouds}. Pillarization can be seen as a special type of 3D voxelization with only one layer of voxels vertically. We annotate each point's global position $[x, y, z]$ with its distance to the LiDAR origin $r$ and its position relative to the voxel's center at $[x_c, y_c, z_c]$. The resultant geometric feature can be expressed as a 7-dimensional vector: $[x, y, z, r, x-x_c, y-y_c, z-z_c]$. Optionally, a simple one-layer fully connected network (similar to the design of like VoxelNet~\cite{Zhou2018VoxelNet:Detection}) can be applied to each voxel to extract more local features.


\subsection{Semantic and Geometric Feature Aggregation}
We use concatenation as the method of aggregating semantic features and geometric features together. The embeddings of those points that are not assigned with semantic features (discarded during range image projection)  are padded with zeros. For each voxel, a locally aggregated feature vector is generated from point-wise embeddings via symmetric pooling operations: we apply element-wise max pooling on high-dimensional semantic features and average pooling on low-dimensional geometric features. Now, we are able to obtain a voxel-wise feature vector that encodes both semantic and geometric features of the set of points inside the voxel. 

\subsection{Temporal Aggregation}

When multi-frame data is available, we add a timestamp $t$ as an additional channel to each point. Such temporal aggregation requires a new design of range image projection for semantic feature extractor. For example, when we aggregate 10 consecutive sweeps, the point cloud is now 10 times denser and thus a large portion of points will be discarded by the pseudo range image projection process. To prevent such loss of information, we propose two solutions: temporal multi-frame fusion and spatial multi-frame fusion. The ablation study of these two aggregation methods is discussed in Section~\ref{sec:temp_aggr}.

\subsubsection{Temporal Multi-Frame Fusion}
Temporal multi-frame fusion retrieves the pseudo range image at each frame respectively, and then concatenates them along a new dimension to form a batch of images as input. This is equivalent to running the same feature extractor on each individual frame.

\subsubsection{Spatial Multi-Frame Fusion}
In spatial multi-frame fusion, we transform all frames' points to the keyframe's coordinate system and increase the resolution of the pseudo range image to allow more points to be projected. The main design choice required here is the multiple $n$ between the new linear resolution and the single-frame resolution. Fig. \ref{fig:occupancy_map} shows the occupancy map of different $n$. When $n$ is too large, the range image becomes sparse and inefficient for dense feature extraction. Ideally, we want the occupancy rate $\tau$ to be as close as possible to the original one. For NuScenes dataset (20 Hz frame rate), we choose $n=2$ for 10-frame aggregation. Notice that in this setting, the range image has only $4\times$ pixels while the 3D point cloud has $10\times$ points. When multiple points are projected to the same pixel, we prioritize those with timestamps closer to the key-frame. Spatial multi-frame fusion allows us to enhance the resolution of input range image efficiently without too much redundancy caused by close or repeating points.

\begin{table*}[t!]
    \begin{center}
    \resizebox{\textwidth}{!}{\begin{tabular}{|l |c c c c c c c c c c| c |}
    \hline
    & car &  truck & bus & trailer &  cons.  &  pedes. &  mcycle & bicycle & cone & barrier & mAP  \\
     \hline \hline
    Point Pillars~\cite{Lang2018PointPillars:Clouds} & 68.4 & 23.0 & 28.2 & 23.4 & 4.1 & 59.7 & 27.4 & 1.1& 30.8 & 38.9 & 30.5 \\
    SARPNET~\cite{Ye2020SARPNET:Detection} & 59.9 & 18.7 & 19.4 & 18.0 & 11.6 & 69.4 & 29.8 & 14.2& 44.6 & 38.3 & 32.4 \\
    CBGS ~\cite{Zhu2019Class-balancedDetection} & \textbf{81.1} & \textbf{48.5} & \textbf{54.9} & 42.9 & 10.5 & \textbf{80.1} & 51.5 & 22.3 & 70.9 & \textbf{65.7} & 52.8 \\
    Ours & 80.1 & 45.4 & 54.0 & \textbf{51.6} & \textbf{15.1} & 79.1 & \textbf{53.1} & \textbf{31.3} & \textbf{71.8} & 62.9 & \textbf{54.5} \\

    \hline
    \end{tabular}}
    \end{center}
    \caption[Detection mAP by categories compared on NuScenes test set]{Detection mAP by categories compared on NuScenes test set.}
    \label{tab:map_by_cat}
\end{table*}

\begin{table*}[t!]
    \begin{center}
    \resizebox{\textwidth}{!}{\begin{tabular}{|l |c c c c c c c c c c| c |}
    \hline 
     & car &  truck & bus & trailer &  cons.  &  pedes. &  mcycle & bicycle & cone & barrier & mAP \\
     \hline \hline
    CBGS*~\cite{Zhu2019Class-balancedDetection}  & 79.8 & 45.8 & 58.6 & 31.1 & 11.7 & 74.8 & 38.3 & 14.2 & 55.0 & 56.6 & 46.6 \\
    Ours w/o sem. feat & 80.1 & 44.2 & 59.1 & 32.2 & 10.9 & 74.5 & 40.2 & 20.2 & 57.8 & 55.6 & 47.5 \\
    Ours & \textbf{82.6} & \textbf{49.9} & \textbf{62.4} & \textbf{36.3} & \textbf{11.8} & \textbf{80.6} & \textbf{53.8} & \textbf{33.8} & \textbf{67.2} & \textbf{64.5} & \textbf{54.3} \\

    \hline
    \end{tabular}}
    \end{center}
    \caption[Detection mAP by categories compared on NuScenes validation set]{Detection mAP by categories compared on NuScenes validation set. *: reproduced with officially released code and our experimental setup. The second line shows the result of our model without aggregation of range-image-based semantic features (row i. in Tab. \ref{tab:abl_study}).}
    \label{tab:map_by_cat_val}
\end{table*}
\subsection{Detector}
As discussed in Section \ref{secVoxel}, the input of the detector can have two types of formats: 2D pillars or 3D voxels. The detector is designed accordingly. For 2D pillar input, we can directly apply an SFPN as the backbone to get the final feature map. On the other hand, for 3D voxel input with the shape of $[H, W, D, C]$, we first adopt a sparse 3D ResNet to downscale the tensor to $[H/s_H, W/s_W, D/s_D, C]$, where $s_H, s_W, s_D$ are the downscale factors. Then we lower the dimension of the tensor by reshaping it to $[H/s_H, W/s_W, D\times C/s_D]$, so that it can be similarly fed into a 2D RPN to generate the BEV feature map. The bounding box regression head is attached to the feature map. We follow the multi-group head design as in CBGS \cite{Zhu2019Class-balancedDetection}. The detailed detector structure is illustrated in Fig.~\ref{fig:panonet3d_detector}.


\subsection{Data Augmentation}
We make several improvements on data augmentation schematics used in SECOND~\cite{Yan2018Second:Detection}. Ground truth boxes are cropped and saved offline, and then pasted onto the current frame's ground plane. Additionally, we allow the augmented object to randomly rotate around the LiDAR center within 45 degrees (its distance to the center of the frame remains unchanged). We also perform global augmentations that randomly transform the whole point cloud, including translation (within [-0.2m, 0,2m]), rotation (within [-45\degree, 45\degree]) and scaling (within [0.95x, 1.05x]). 

Newly pasted objects may occlude with other objects. Traditional methods often ignore such occlusions, and keep all augmented objects even they are not detectable by a real LiDAR sensor. However, our method naturally solves this problem by removing all annotations that have less than 3 points projected on the pseudo range image. As a result, the objects that should not be visible to the LiDAR are easily filtered out.
\subsection{Implementation Details}
Our implementation is based on CBGS's~\cite{Zhu2019Class-balancedDetection} official code base\footnote{\url{https://github.com/poodarchu/Det3D}}. All object classes share the detection backbone except an exclusive two-layer regression head for each category group. The experiments are conducted on 4 NVIDIA 1080 Ti with PyTorch's official implementation of multi-GPU synchronized batch normalization. We train the network with Adam optimizer~\cite{Kingma2015Adam:Optimization}, one-cycle policy~\cite{AnotherPolicy} (max learning rate: 0.0001, division factor: 5), and the batch size of 4 for 20 epochs. The IoU threshold of the non-maximum suppression is 0.2 and the maximum number of final predicted bounding boxes is 100. The anchors selected as the mean values of all labels. On 1080 Ti, our model runs at 20 fps during inference.
\section{Results}
\begin{table*}[t!]
    \begin{center}
    \resizebox{\textwidth}{!}{\begin{tabular}{| c | l | c c c c | c |}
    \hline
    & Method & Range image feat. & \# Input frames & Voxelization & BEV resolution(m) & mAP \\
     \hline    \hline
    a. & Point Pillars~\cite{Lang2018PointPillars:Clouds} & - & 1 & Pillar &  0.25 & 24.0 \\ 
    b. & Point Pillars~\cite{Lang2018PointPillars:Clouds} & - & 10 & Pillar &  0.25 & 29.5 \\
    c. & Ours                                             & \xmark & 1 & Pillar &  0.25 & 31.5 \\
    d. & CGBS~\cite{Zhu2019Class-balancedDetection}       & - & 1 & Voxel &  0.1 & 39.2 \\
    e. & Ours                                             & \cmark & 1 & Voxel &  0.1 & 43.1 \\
    f. & Ours                                             & \cmark & 1 & Pillar &  0.25 & 45.2 \\
    g. & Ours                                             & \xmark & 10 & Voxel &  0.1 & 46.3 \\
    h. & CGBS~\cite{Zhu2019Class-balancedDetection}       & - & 10 & Voxel &  0.1 & 46.6 \\
    i. & Ours                                             & \xmark & 10 & Voxel &  0.05 & 47.5 \\
    j. & Ours                                             & \cmark & 10 & Pillar &  0.25 & 47.9 \\
    k. & Ours                                             & \cmark & 10 & Pillar &  0.1 & 48.0 \\
    l. & Ours                                             & \cmark & 10 & Voxel &  0.1 & 52.9 \\
    m. & Ours                                             & \cmark & 10 & Voxel &  0.05 & 54.3 \\
    \hline
    \end{tabular}}
    \end{center}
    \caption[Ablation studies on NuScenes validation set]{Ablation studies on NuScenes validation set. `range image feat.' means whether the detector uses perspective-view-based semantic feature extractor. `BEV resolution' means the x-y resolution when the point cloud is voxelized.}
    \label{tab:abl_study}
\end{table*}

\begin{figure*}[h!]
\centering
        \includegraphics[width=0.33\linewidth]{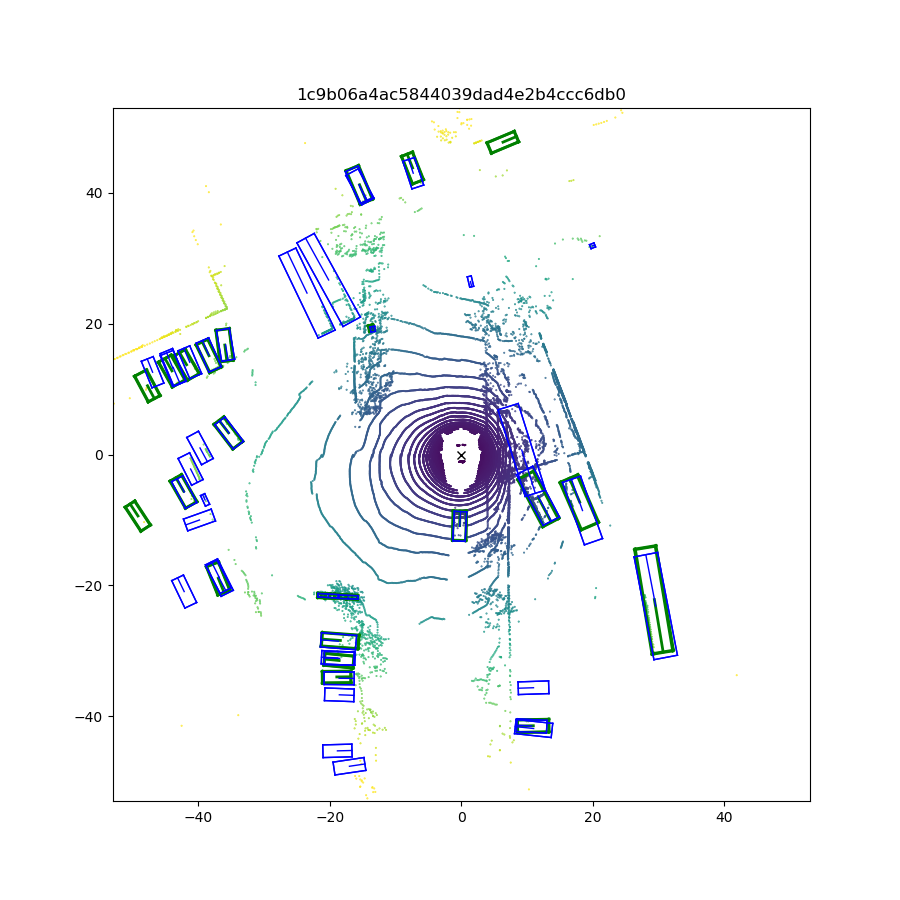}\includegraphics[width=0.33\linewidth]{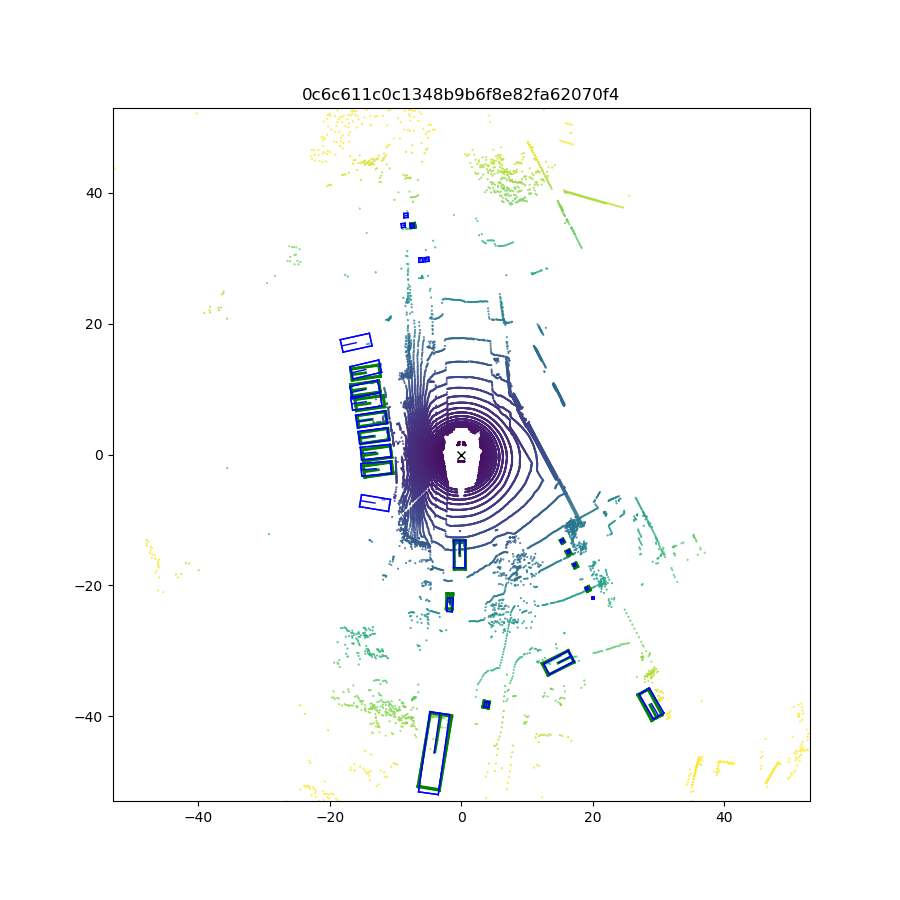}\includegraphics[width=0.33\linewidth]{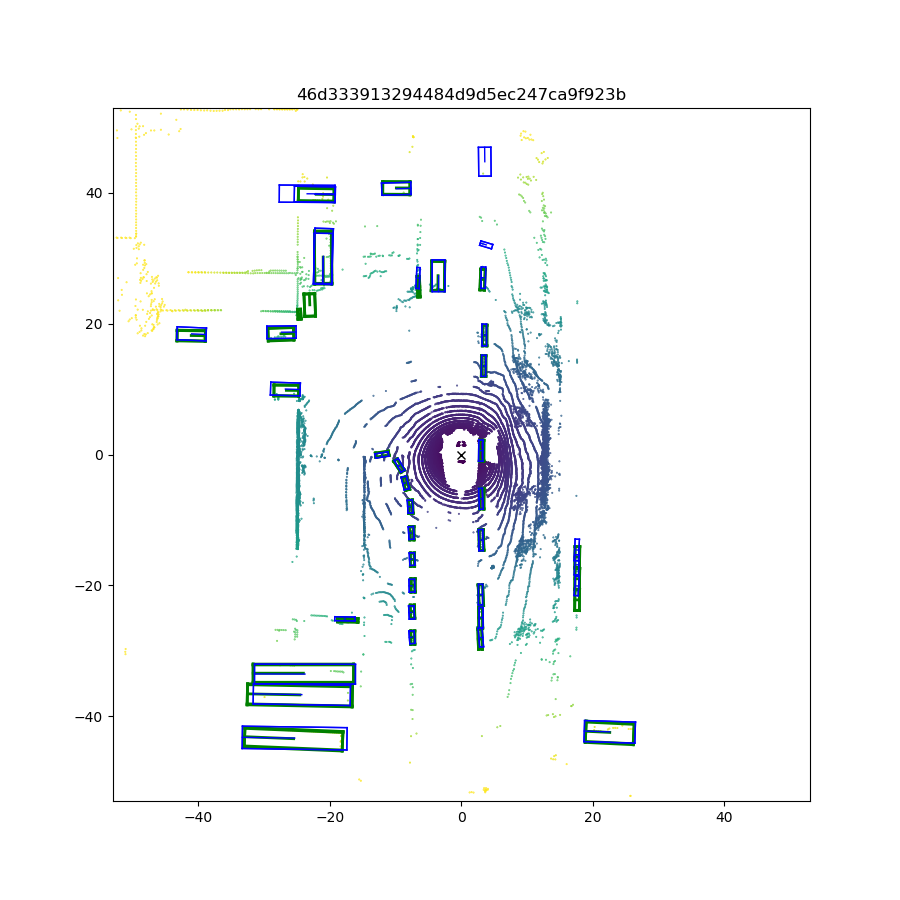}
        \includegraphics[width=0.33\linewidth]{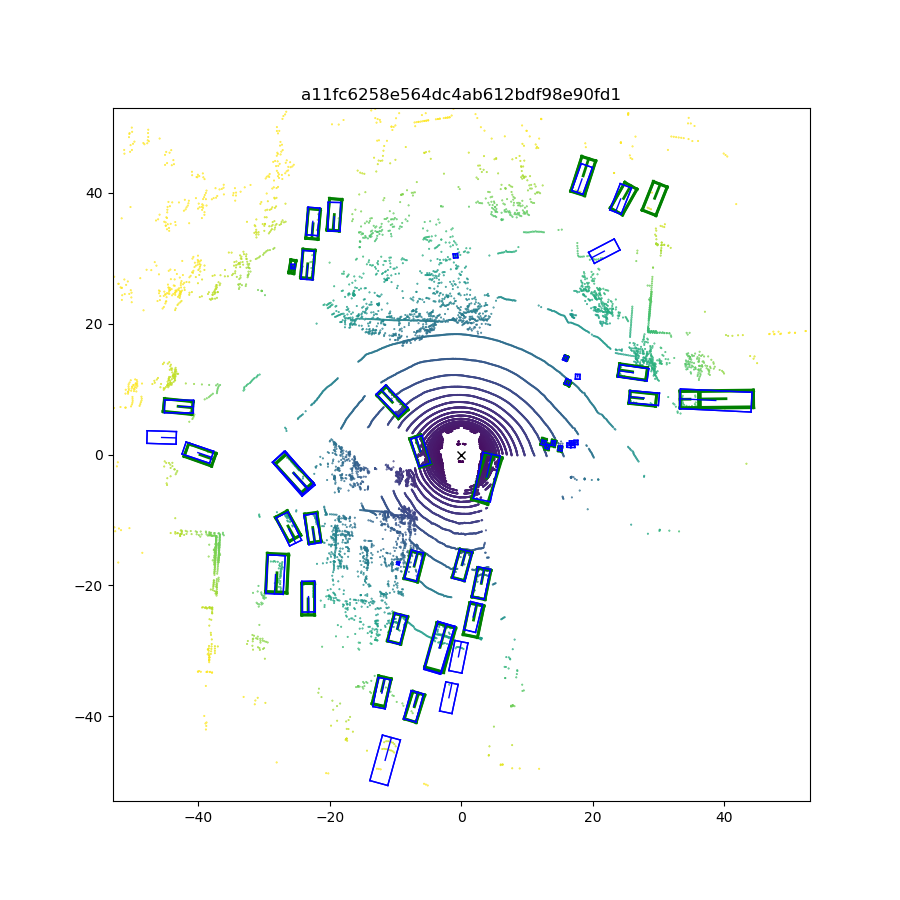}\includegraphics[width=0.33\linewidth]{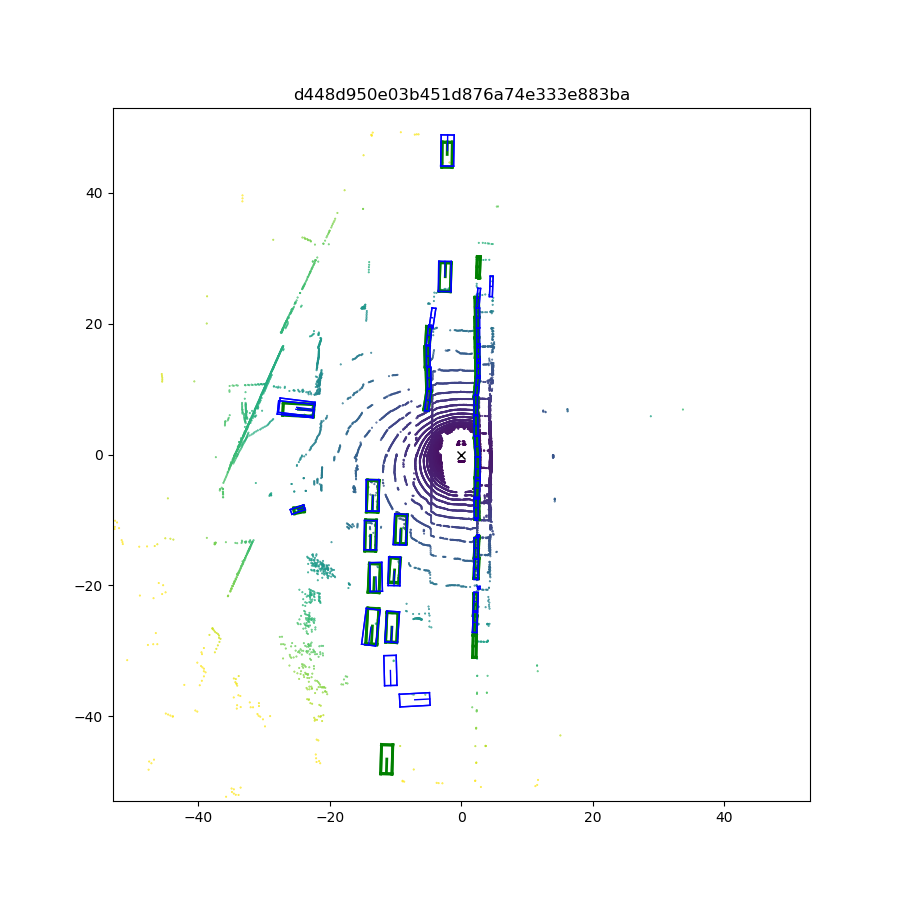}\includegraphics[width=0.33\linewidth]{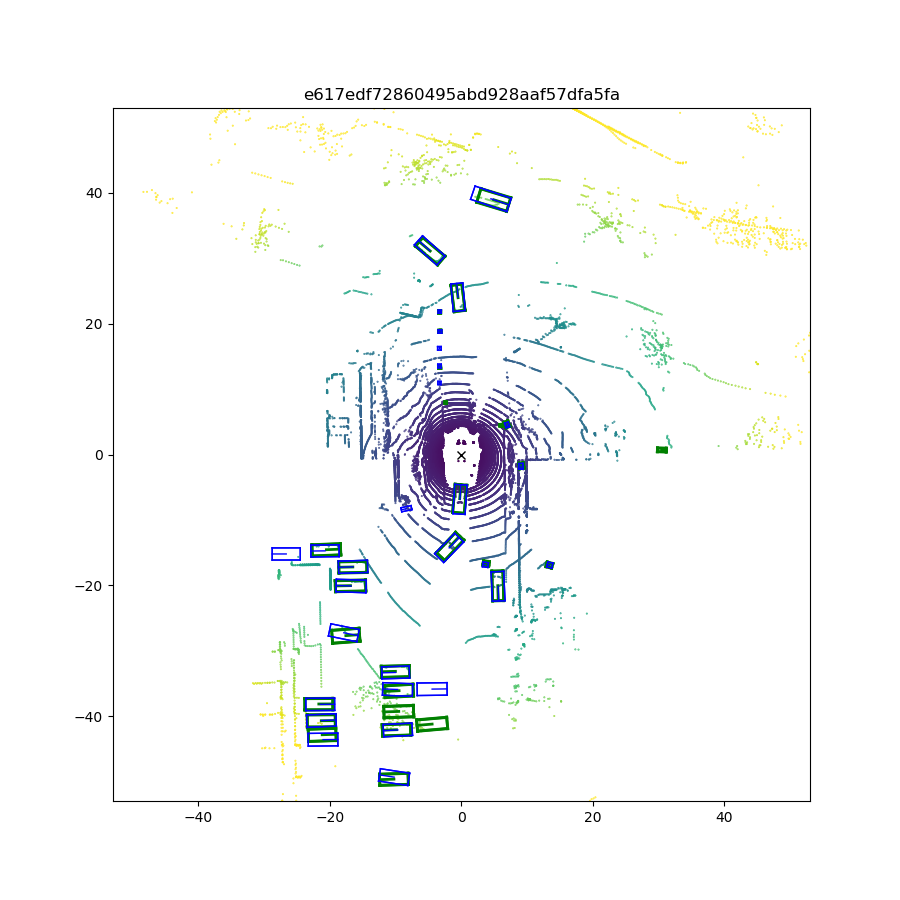}
        \label{fig:panonet3d_results}
        \caption[Detection examples of PanoNet3D on NuScenes dataset]{Detection examples of PanoNet3D on NuScenes dataset. Ground truths are annotated in green boxes and detection results are annotated in blue boxes.}
\end{figure*}

We first compare the quantitative performance of our method against other SOTA methods on the NuScenes dataset, while the results on the KITTI dataset are presented in the supplemental materials. Qualitative results (visualization of predictions) are shown in Fig. \ref{fig:panonet3d_results}. Next, we conduct ablation studies to explain how we make the decisions during network design and show where the performance improvements come from.

\subsection{Main Results}

We submitted the results of our method to the NuScenes test server. In Tab. \ref{tab:map_by_cat}, we compare PanoNet3D against other methods on the NuScenes detection leaderboard. Our overall mAP surpasses the current first-place method CBGS~\cite{Zhu2019Class-balancedDetection} by 1.7\%. For fairness, we also compare our method against CBGS's reproducible performance on NuScenes validation set in Tab. \ref{tab:map_by_cat_val}. The results of CBGS are reproduced with its official code and under the same experimental setup as ours. Our model improves mAP on all categories including 2.5\% on car. Higher performance gains are observed on `tall-and-thin' object categories such as bicycle and cone. These objects have larger projection sizes on depth image rather than on the BEV plane, so our model can achieve better overall understandings compared to traditional detectors. We also re-trained our model and baseline (CBGS) from scratch for 4 more times with different random seeds. The performance errors of all trails are within 0.4\% mAP. 

\subsection{Ablation Study}
Tab. \ref{tab:abl_study} shows a series of ablation studies. Based on these results, we can make the following key observations. Each line of the result is represented with small letters. Across all factors, we find that the pano-view semantic feature extractor contributes the most to the performance gain.

\subsubsection{Baseline Comparison}

The major difference between PanoNet3D and other detectors is its range-image-based semantic feature extractor. Without the aggregation of extracted semantic features, our pillar-based detector should have a similar framework to PointPillar's~\cite{Lang2018PointPillars:Clouds} except for the backbone design. Our pillar-based baseline model achieves better performance than PointPillars (a.-c.). The most likely reason is our SFPN backbone is able to utilize multi-level features more efficiently. On the other hand, for voxel-based detectors, CBGS has similar performance to our baseline model (without the semantic feature extractor), showing that our improvements against CBGS do not come from the different detector backbones used by PanoNet3D and CBGS (g.-h.).

\subsubsection{Range Image Semantic Feature Extractor} For single-frame pillar-based detectors, semantic features extracted from range images significantly improve the average mAP by 13.7\% (c.-f.). With the help of a semantic feature extractor, our single-frame model is able to achieve comparable results against other multi-frame models. For multi-frame voxel-based detectors, semantic features also improve the average mAP by over 6\% (g.-l., i.-m.).  From Tab. \ref{tab:map_by_cat_val}, we can further observe that combining deep semantic features with raw geometric features leads to improvements across all 10 categories. The perspective view is natural to LiDAR sensors and contains semantics that cannot be extracted from real-world Euclidean space, which helps the detector to achieve a better overall understanding of the scene.

\subsubsection{Pillar or Voxel} For single-frame input, the pillar-based detector performs slightly better (e.-f.), while for multi-frame input, the voxel-based detector is more favorable (k.-l.). One possible explanation is that the pillar-based detector is sufficient for the density of a single-frame point cloud and can prevent over-fitting caused by more complex 3D convolutions. Multi-frame input has much denser point clouds whose features can not be well extracted by the simpler pillar feature extractor. 

\subsubsection{BEV Resolution} Finer grids of voxelization usually lead to better detection performance. However, its impact is less dominant than other factors. Increasing BEV resolution from 0.25m to 0.1m improves mAP of 10-frame pillar-based detector by 0.1\% (j,-k.), and increasing BEV resolution from 0.1m to 0.05m improves mAP of 10-frame voxel-based detector by 1.4\% (l.-m.). 

\subsection{Voxel Feature Pooling Methods}
We test all pooling methods during aggregating point-wise features to voxel-wise features. With all combinations shown in Tab. \ref{tab:feature_aggregation}, we conclude that max pooling on higher-dimensional semantic features with average pooling on lower-dimensional geometric features yields the best results.
\begin{table}[h!]
    \begin{center}
    \begin{tabular}{| c c | c |}
    \hline
    Deep sem. feat. aggr. & Raw geo. feat. aggr. & mAP \\
     \hline
    Max & Max & 44.6 \\
    Max & Average & 45.2 \\
    Average & Max & 44.3 \\
    Average & Average & 44.9 \\
    \hline
    \end{tabular}
    \end{center}
    \caption[Study on pooling methods during voxel-wise feature aggregation]{Study on pooling methods during voxel-wise feature aggregation. The experiment is done with a single-frame pillar detector as baseline.}
    \label{tab:feature_aggregation}
\end{table}

\subsection{Temporal Aggregation Methods}\label{sec:temp_aggr}
We also experiment with different temporal aggregation approaches, the results of which are shown in Tab. \ref{tab:temp_aggregation}. Spatial multi-frame fusion with $n=2$ is the best among them, showing that our previous analysis is correct. However, notice that the optimal $n$ is not a fixed value. If the number of aggregated frames or the input dataset changes, we might need to change $n$ accordingly to accommodate the data. 
\begin{table}[h!]
    \begin{center}
    \begin{tabular}{| c | c | c |}
    \hline
    Aggregation method  & $n$ & mAP \\
     \hline \hline
    Temporal 10-frame fusion & - & 52.9 \\
    Spatial 10-frame fusion & 1 & 53.1 \\
    Spatial 10-frame fusion & 2 & 54.3 \\
    Spatial 10-frame fusion & 3 & 52.2 \\
    \hline
    \end{tabular}
    \end{center}
    \caption[Results of different multi-frame temporal aggregation approaches]{Results of different multi-frame temporal aggregation approaches.}
    \label{tab:temp_aggregation}
\end{table}

\section{Conclusion} \label{chapConclusions}

We explore the possibility of combining both semantic and geometric understanding of 3D LiDAR point clouds. Experiments show that both objects' appearance to the sensor and their actual shapes in 3D space are important for detection networks.  By enhancing each point's raw geometric coordinates with deep semantic features extracted from pseudo range images, we are able to achieve a better understanding of the scene and better overall detection performance.

{\small
\bibliographystyle{ieee}
\bibliography{references}
}

\end{document}